## Valuation-Based Systems for Discrete Optimization

by
Prakash P. Shenoy
*School of Business, University of Kansas, Lawrence, KS 66045-2003, USA*
*Tel: (913)-864-7551, Fax: (913)-864-5328, Bitnet: PSHENOY@UKANVM*

### Abstract

This paper describes valuation-based systems for representing and solving discrete optimization problems. The information in an optimization problem is represented using variables, sample spaces of variables, a set of values, and functions that map sample spaces of sets of variables to the set of values. The functions, called valuations, represent the factors of an objective function. Solving the optimization problem involves using two operations called combination and marginalization. The combination operation tells us how to combine the factors of the objective function to form the global objective function. Marginalization is either maximization or minimization. Solving an optimization problem can be simply described as finding the marginal of the joint objective function for the empty set. We state some simple axioms that combination and marginalization need to satisfy to enable us to solve an optimization problem using local computation.

### 1. Introduction

The main objective of this paper is to describe a valuation-based system (VBS) for representing and solving discrete optimization problems. There are several reasons why this is useful.

First, I initially proposed VBSs for managing uncertainty in expert systems [Shenoy, 1989a, 1989b]. Here I show that these systems also have the expressive power to represent and solve optimization problems.

Second, problems in decision analysis involve managing uncertainty and optimization. That both of these problems can be solved in a common framework suggests that decision problems also can be represented and solved in the framework of VBS. Indeed, Shenoy [1990a] shows that this is true. In fact, the solution procedure for VBS when applied to decision problems results in a method that is computationally more efficient than decision trees and influence diagrams.

Third, the solution procedure of VBS when applied to optimization problems results in a method called non-serial dynamic programming [Bellman, 1957; Bertele and Brioschi, 1972]. Thus in an abstract sense, the local computation algorithms that have been described by Pearl [1986], Shenoy and Shafer [1986], Dempster and Kong [1988], and Lauritzen and Spiegelhalter [1988] are just dynamic programming.

Fourth, in this paper we describe some simple axioms for combination and marginalization that enable the use of dynamic programming for solving optimization problems. We believe these axioms are new. They are weaker than those proposed by Mitten [1962].

Fifth, the VBS described here can be easily adapted to represent propositional logic [Shenoy 1989a, 1990b] and constraint satisfaction problems [Shenoy and Shafer, 1988b].

An outline of this paper is as follows. In Section 2, we show how to represent an optimization problem as a VBS. In Section 3, we state some simple axioms that justify the use of local computation in solving VBSs. In Section 4, we show how to solve a VBS. Throughout the paper, we use one example to illustrate all definitions and the solution process.

### 2. Representation of Optimization Problems

A valuation-based representation of an optimization problem uses variables, frames, and valuations. We will discuss each of these in detail. We will illustrate all definitions using an optimization problem from Bertele and Brioschi [1972].

**An Optimization Problem.** There are five variables labeled as A, B, C, D, and E. Each variable has two possible values. Let a and ~a denote the possible values of A, etc. The global objective function F for variables A, B, C, D, and E factors additively as follows: $F(v,w,x,y,z) =$

$F_1(v,x,z) + F_2(v,w) + F_3(w,y,z)$, where $F_1$, $F_2$, and $F_3$, are as shown in Figure 1 below. The problem is to find the minimum value of F and a configuration $(v,w,x,y,z)$ that minimizes F.

**Figure 1.** The factors of the objective function, $F_1$, $F_2$, and $F_3$.

| $w \in \mathcal{W}_{\{A,C,E\}}$ | | | $F_1(w)$ |
|---|---|---|---|
| a | c | e | 1 |
| a | c | ~e | 3 |
| a | ~c | e | 5 |
| a | ~c | ~e | 8 |
| ~a | c | e | 2 |
| ~a | c | ~e | 6 |
| ~a | ~c | e | 2 |
| ~a | ~c | ~e | 4 |

| $w \in \mathcal{W}_{\{A,B\}}$ | | $F_2(w)$ |
|---|---|---|
| a | b | 4 |
| a | ~b | 8 |
| ~a | b | 0 |
| ~a | ~b | 5 |

| $w \in \mathcal{W}_{\{B,D,E\}}$ | | | $F_3(w)$ |
|---|---|---|---|
| b | d | e | 0 |
| b | d | ~e | 5 |
| b | ~d | e | 6 |
| b | ~d | ~e | 3 |
| ~b | d | e | 5 |
| ~b | d | ~e | 1 |
| ~b | ~d | e | 4 |
| ~b | ~d | ~e | 3 |

**Variables and Configurations.** We use the symbol $\mathcal{W}_X$ for the set of possible values of a variable X, and we call $\mathcal{W}_X$ the *frame* for X. We will be concerned with a finite set $\mathfrak{X}$ of variables, and we will assume that all the variables in $\mathfrak{X}$ have finite frames.

Given a finite non-empty set h of variables, we let $\mathcal{W}_h$ denote the Cartesian product of $\mathcal{W}_X$ for X in h, i.e., $\mathcal{W}_h = \times \{ \mathcal{W}_X \mid X \in h \}$. We call $\mathcal{W}_h$ the *frame* for h. We call elements of $\mathcal{W}_h$ *configurations of h*. Lower-case bold-faced letters, such as x, y, etc., will denote configurations. If x is a configuration of g, y is a configuration of h, and $g \cap h = \varnothing$, then $(x,y)$ will denote the configuration of $g \cup h$ obtained by concatenating x and y.

It will be convenient to allow the set of variables h to be empty. We will adopt the convention that the frame for the empty set $\varnothing$ consists of a single element, and we will use the symbol $\blacklozenge$ to name that element; $\mathcal{W}_\varnothing = \{ \blacklozenge \}$. If x is a configuration of g, then $(x, \blacklozenge)$ is simply x.

**Values and Valuations.** We will be concerned with a set $\mathbb{V}$ whose elements are called *values*. $\mathbb{V}$ may be finite or infinite. Given a set h of variables, we call any function $H: \mathcal{W}_h \to \mathbb{V}$, a *valuation for h*. Note that to specify a valuation for $\varnothing$, we need to specify only a single value, the value of $H(\blacklozenge)$.

In our problem, the set $\mathbb{V}$ corresponds to the set of real numbers, and we have three valuations $F_1, F_2$ and $F_3$. $F_1$ is a valuation for $\{A,C,E\}$, $F_2$ is a valuation for $\{A,B\}$ and $F_3$ is a valuation for $\{B,D,E\}$. Figure 2 shows a graphical depiction of the optimization problem, called a valuation network. In a valuation network, variables are shown as squares and valuations are shown as diamonds. Each valuation is linked to the variables it is defined for.

We will let $\mathcal{V}_h$ denote the set of all valuations for h, and $\mathcal{V}$ denote the set of all valuations, i.e., $\mathcal{V} = \cup \{ \mathcal{V}_h \mid h \subseteq \mathfrak{X} \}$.

**Projection and Extension of Configurations.** *Projection* of configurations simply means dropping extra coordinates; if $(\sim a,b,\sim c,d,e)$ is a configuration of $\{A,B,C,D,E\}$, for example, then the projection of $(\sim a,b,\sim c,d,e)$ to $\{A,C,E\}$ is simply $(\sim a,\sim c,e)$, which is a configuration of $\{A,C,E\}$.

If g and h are sets of variables, $h \subseteq g$, and x is a configuration of g, then we will let $x^{\downarrow h}$ denote the projection of x to h. The projection $x^{\downarrow h}$ is always a configuration of h. If h=g and x is a configuration of g, then $x^{\downarrow h} = x$. If $h = \varnothing$, then of course $x^{\downarrow h} = \blacklozenge$.

**Figure 2.** The valuation network for the optimization problem.

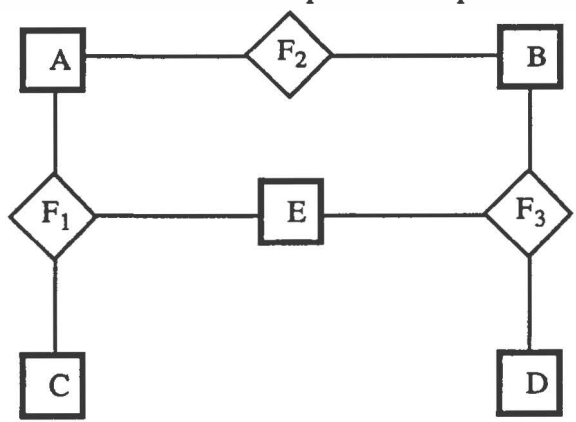

**Combination**. We assume there is a mapping $\copyright:\mathbb{V}\times\mathbb{V}\rightarrow\mathbb{V}$ called *combination* so that if $u, v \in \mathbb{V}$, then $u\copyright v$ is the value representing the combination of u and v. We define a mapping $\oplus:\mathcal{V}\times\mathcal{V}\rightarrow\mathcal{V}$ in terms of $\copyright$, also called *combination*, such that if G and H are valuations for g and h respectively, then $G\oplus H$ is the valuation for $g\cup h$ given by

$$(G\oplus H)(\mathbf{x}) = G(\mathbf{x}^{\downarrow g})\copyright H(\mathbf{x}^{\downarrow h}) \tag{2.1}$$

for all $\mathbf{x}\in \mathcal{W}_g$. We call $G\oplus H$ the *combination of G and H*.

In our optimization problem, $\copyright$ is simply addition. We can express the global objective function F as follows $F = F_1\oplus F_2\oplus F_3$.

**Marginalization**. We assume that for each $h\subseteq\mathfrak{X}$, there is a mapping

$\downarrow h:\cup\{\mathcal{V}_g \mid g\supseteq h\}\rightarrow\mathcal{V}_h$, called *marginalization to h*, such that if G is a valuation for g and $g\supseteq h$, then $G^{\downarrow h}$ is a valuation for h. We call $G^{\downarrow h}$ the *marginal of G for h*.

For our optimization problem, we define marginalization as follows:

$$G^{\downarrow h}(\mathbf{x}) = \text{MIN}\{G(\mathbf{x},\mathbf{y}) \mid \mathbf{y}\in \mathcal{W}_{g\text{-}h}\} \tag{2.2}$$

for all $\mathbf{x}\in \mathcal{W}_h$. Thus, if F is an objective function, then $F^{\downarrow\varnothing}(\blacklozenge)$ represents the minimum value of F.

In an optimization problem, besides the minimum value, we are usually also interested in finding a configuration where the minimum of the joint valuation is achieved. This motivates the following definition.

**Solution for a Valuation**. Suppose H is a valuation for h. We will call $\mathbf{x}\in \mathcal{W}_h$ a *solution for H* if $H(\mathbf{x}) = H^{\downarrow\varnothing}(\blacklozenge)$.

**Solution for a Variable**. As we shall see, computing a solution for a valuation is a matter of bookkeeping. Each time we eliminate a variable from a valuation using minimization, we store a table of configurations of the eliminated variable where the minimums are achieved. We can think of this table as a function. We call this function "a solution for the variable." Formally, we define a solution for a variable as follows. Suppose X is a variable, suppose h is a subset of variables containing X, and suppose H is a valuation for h. We call a function $\Psi_X: \mathcal{W}_{h-\{X\}} \rightarrow \mathcal{W}_X$ a *solution for X (with respect to H)* if

$$H^{\downarrow(h-\{X\})}(\mathbf{c}) = H(\mathbf{c},\Psi_X(\mathbf{c})) \tag{2.3}$$

for all $\mathbf{c}\in \mathcal{W}_{h-\{X\}}$.

If $\mathfrak{X}$ is a large set of variables, then a brute force computation of F and an exhaustive search of the set of all configurations of $\mathfrak{X}$ to determine a solution for F is not possible. In the next section we will state axioms for combination and marginalization that make it possible to use local computation to compute the minimum value of F and a solution for F.

## 3. The Axioms

We will list three axioms. Axiom A1 is for combination. Axiom A2 is for marginalization. And Axiom A3 is for combination and marginalization.

**A1** (*Commutativity and associativity of combination*): Suppose u, v, w are values. Then $u \otimes w = v \otimes u$ and $u \otimes (v \otimes w) = (u \otimes v) \otimes w$.

**A2** (*Consonance of marginalization*): Suppose G is a valuation for g, and $k \subseteq h \subseteq g$. Then $(G^{\downarrow h})^{\downarrow k} = G^{\downarrow k}$.

**A3** (*Distributivity of marginalization over combination*): Suppose G and H are valuations for g and h, respectively. Then $(G \oplus H)^{\downarrow g} = G \oplus (H^{\downarrow g \cap h})$.

It follows from axiom A1 that $\oplus$ is commutative and associative. Therefore, the combination of several valuations can be written without using parentheses. For example, $(...((F_1 \oplus F_2) \oplus F_3) \oplus ... \oplus F_k)$ can be simply written as $\oplus \{F_i, ..., F_k\}$ without indicating the order in which to do the combination.

If we regard marginalization as a reduction of a valuation by deleting variables, then axiom A2 can be interpreted as saying that the order in which we delete the variables does not matter.

Axiom A3 is the crucial axiom that makes local computation of marginals and solution possible. Axiom A3 states that computation of $(G \oplus H)^{\downarrow g}$ can be done without having to compute $G \oplus H$.

## 4. Solving a VBS Using Local Computation

Suppose we are given a collection of valuations $\{F_1, ..., F_k\}$ where each valuation $F_i$ is for subset $h_i$ of $\mathfrak{X}$. The problem is (i) to find the minimum value of $F = \oplus \{F_1, ..., F_k\}$ and (ii) to find a solution for F. We will assume that combination and marginalization satisfy the axioms.

We will call the collection of subsets $\{h_1, ..., h_k\}$ for which we have valuations a *hypergraph* and denote it by $\mathfrak{H}$.

Solving a VBS proceeds in three phases. In phase one, we arrange the subsets of variables in $\mathfrak{H}$ in a "rooted Markov tree." In the phase two, we "propagate" the valuations $\{F_1, ..., F_n\}$ in the rooted Markov tree using a local message-passing scheme resulting in the computation of the marginal $F^{\downarrow \varnothing}$. In the phase three, we construct a solution for F again using a local message-passing scheme.

### 4.1. Phase One: Finding a Rooted Markov Tree Arrangement

A *Markov tree* is a topological tree, whose vertices are subsets of variables, with the property that when a variable belongs to two distinct vertices, then every vertex lying on the path between these two vertices contains the variable.

A *rooted Markov tree* is a Markov tree with the empty subset $\varnothing$ as the root and such that all edges in the tree are directed toward the root.

First, note that the only information we need in phase one is the set $\mathfrak{H}$. Second, in arranging a set of subsets in a rooted Markov tree, we may have to add some subsets to the hypergraph $\mathfrak{H}$. Third, in general, there may be many rooted Markov tree arrangements of a hypergraph. Figure 3 shows a rooted Markov tree arrangement of the subsets {A,C,E}, {A,B}, and {B,D,E}. Subsets {A,E}, {B,E}, {A,B,E}, {A}, and $\varnothing$ are added during the arrangement process.

The computational efficiency of phase two depends on the sizes of the frames of the vertices of the Markov tree constructed in the phase one. Finding an optimal rooted Markov tree (a rooted Markov tree whose largest frame is as small as possible) has been shown to be a NP-complete problem [Arnborg *et al.*, 1987]. Thus we have to balance the computational efforts in the two phases. We should emphasize, however, that this is strictly a computational effort question. If computational effort is not an issue, then it does not matter which rooted Markov tree is used for propagating the valuations. All rooted Markov trees give the same final answer, i.e., the marginal of the joint valuation for the empty set. We will describe a heuristic called "one-step-look-ahead" due to Kong [1986] to find a good rooted Markov tree.

---

**Figure 3**. A rooted Markov tree for the optimization problem.

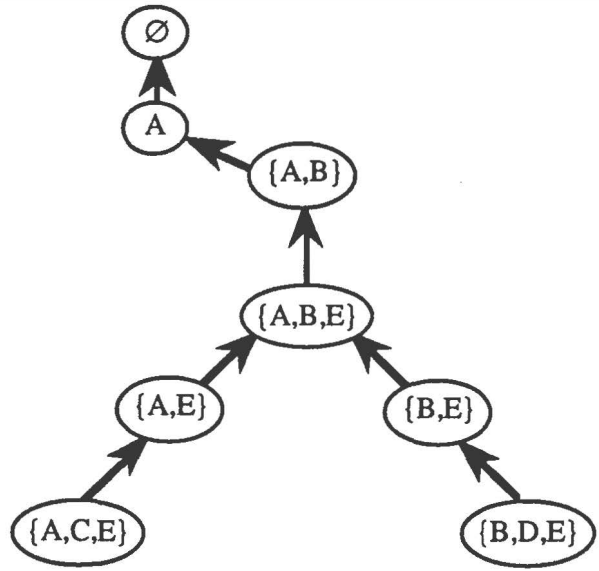

---

The method described below for arranging a hypergraph in a rooted Markov tree is due to Kong [1986] and Mellouli [1987].

Suppose $\mathfrak{H}$ is a hypergraph on $\mathfrak{X}$. To arrange the subsets in $\mathfrak{H}$ in a Markov tree, we first pick a sequence of variables in $\mathfrak{X}$. As we shall see, each sequence of the variables gives rise to a Markov tree arrangement. Mellouli [1987] has shown that an optimal Markov tree arrangement can be found by picking some sequence. Of course, since there are an exponential number of sequences, finding an optimal sequence is, in general, a difficult problem.

Consider the following set of instructions in pseudo-Pascal:

$$u := \mathfrak{X} \quad \{\text{Initialization}\}$$
$$\mathfrak{H}_0 := \mathfrak{H} \quad \{\text{Initialization}\}$$
$$V := \varnothing \quad \{\text{Initialization}\}$$
$$E := \varnothing \quad \{\text{Initialization}\}$$
**for** $i = 1$ **to** n **do**
  **begin**
    Pick a variable from set $u$ and call it $X_i$
    $$u := u - \{X_i\}$$
    $$g_i := \cup\{h \in \mathfrak{H}_{i-1} \mid X_i \in h\}.$$
    $$f_i := g_i - \{X_i\}.$$
    $$V := V \cup \{h \in \mathfrak{H}_{i-1} \mid X_i \in h\} \cup \{f_i\} \cup \{g_i\}$$
    $$E := E \cup \{(h, g_i) \mid h \in \mathfrak{H}_{i-1}, h \neq g_i, X_i \in h\} \cup \{(g_i, f_i)\}$$
    $$\mathfrak{H}_i := \{h \in \mathfrak{H}_{i-1} \mid X_i \notin h\} \cup \{f_i\}$$
  **end** $\{\text{for}\}$

After the execution of the above set of instructions, it is easily seen that the pair $(V, E)$ is a rooted Markov tree arrangement of $\mathcal{H}$ where $V$ denotes the set of vertices of the rooted Markov tree and $E$ denotes the set of edges directed toward the root. Note that at each iteration of the above sequence of instructions, we add subsets $g_i$ and $f_i$ to the set of subsets if they are not already members of $\mathcal{H}$.

We shall say that in the $i^{th}$ iteration of the for-loop in the above set of instructions, the variable $X_i$ that is picked from set u is *marked*. Note that the subsets in $\mathcal{H}_i$ contain only unmarked variables.

Kong [1986] has suggested a heuristic called *one-step-look-ahead* for finding a good Markov tree. This heuristic tells us which variable to mark next. As the name of the heuristic suggests, the variable that should be marked next is an unmarked variable $X_i$ such that the cardinality of $\mathcal{W}_{f_i}$ is the smallest. Thus, the heuristic attempts to keep the sizes of the frames of the added vertices as small as possible by focussing only on the next subset added. In the optimization problem, a sequence selected by the one-step-look-ahead procedure is C,D,E,B,A. Figure 3 shows the resulting rooted Markov tree. See Zhang [1988] for other heuristics for good Markov tree construction.

### 4.2. Phase Two: Finding the Marginal of the Joint Valuation

Suppose we have arranged the hypergraph $\mathcal{H}$ in a rooted Markov tree. Let $\mathcal{H}'$ denote the set of subsets in the Markov tree. Clearly $\mathcal{H}' \supseteq \mathcal{H}$. To simplify the exposition, we will assume that there is exactly one valuation for each non-empty subset $h \in \mathcal{H}'$. If h is a subset that was added during the rooted Markov tree construction process, then we can associate the vacuous valuation (the valuation whose values are all 0) with it. On the other hand, if subset h had more than one valuation defined for it, then we can combine these valuations to obtain one valuation.

First, note that the rooted Markov tree defines a parent-child relation between adjacent vertices. If there is an edge $(h_i, h_j)$ in the rooted Markov tree, we will refer to $h_i$ as $h_j$'s *parent* and refer to $h_j$ as $h_i$'s *child*. Let $h_0 = \varnothing$ denote the *root* of the Markov tree. Let Pa(h) denote h's parent and let Ch(h) denote the set of h's children. Every non-root vertex has exactly one parent. Some vertices have no children and we will refer to such vertices as *leaves*. Note that the root has exactly one child.

In describing the process of finding the marginal of the joint valuation for the empty set, we will pretend that there is a processor at each vertex of the rooted Markov tree. Also, we will assume that these processors are connected using the same architecture as the Markov tree. In other words, each processor can directly communicate only with its parent and its children.

In the propagation process, each subset (except the root $h_0$) transmits a valuation to its parent. We shall refer to the valuation transmitted by subset $h_i$ to its parent Pa($h_i$) as a *valuation message* and denote it by $V^{h_i \to Pa(h_i)}$. Suppose $\mathcal{H}' = \{h_0, h_1, ..., h_k\}$ and let $F_i$ denote the valuation associated with non-empty subset $h_i$. Then, the valuation message transmitted by a subset $h_i$ to its parent Pa($h_i$) is given by

$$V^{h_i \to Pa(h_i)} = (\oplus \{ V^{h \to h_i} \mid h \in Ch(h_i) \} \oplus F_i)^{\downarrow(h_i \cap Pa(h_i))} \tag{4.1}$$

In words, the valuation message transmitted by a subset to its parent consists of the combination of the valuation messages it receives from its children plus its own valuation suitably marginalized. Note that the combination operation that is done in (4.1) is on the frame $\mathcal{W}_{h_i}$.

Expression (4.1) is a recursive formula. We need to start the recursion somewhere. Note that if subset $h_i$ has no children, then Ch($h_i$) = $\varnothing$ and the expression in (4.1) reduces to

$$V^{h_i \to Pa(h_i)} = (F_i)^{\downarrow(h_i \cap Pa(h_i))} \tag{4.2}$$

Thus the leaves of the Markov tree (the subsets that have no children) can send valuation messages to their parents right away. The others wait until they have heard from all their children before they send a valuation message to their parent.

The following theorem states that the valuation message from $h_0$'s child is indeed the desired marginal.

**Theorem 1.** The marginal of the joint valuation for the empty set is equal to the message received by the root, i.e.,

$$(F_1 \oplus ... \oplus F_k)^{\downarrow \varnothing} = V^{h \to h_0}.$$

The essence of the propagation method described above is to combine valuations on smaller frames instead of combining all valuations on the global frame associated with $\mathfrak{X}$. To ensure that this method gives us the correct answers, the smaller frames have to be arranged in a rooted Markov tree.

Figure 4 shows the propagation of valuations in the optimization problem. Figure 5 shows the details of the valuation messages. As is clear from Figure 5, the minimum value of the function F is 2.

**Figure 4**. The propagation of valuations in the optimization problem. The valuation messages are shown as rectangles overlapping the corresponding edges. The valuations associated with the vertices are shown as diamonds linked to the corresponding vertices by dotted lines.

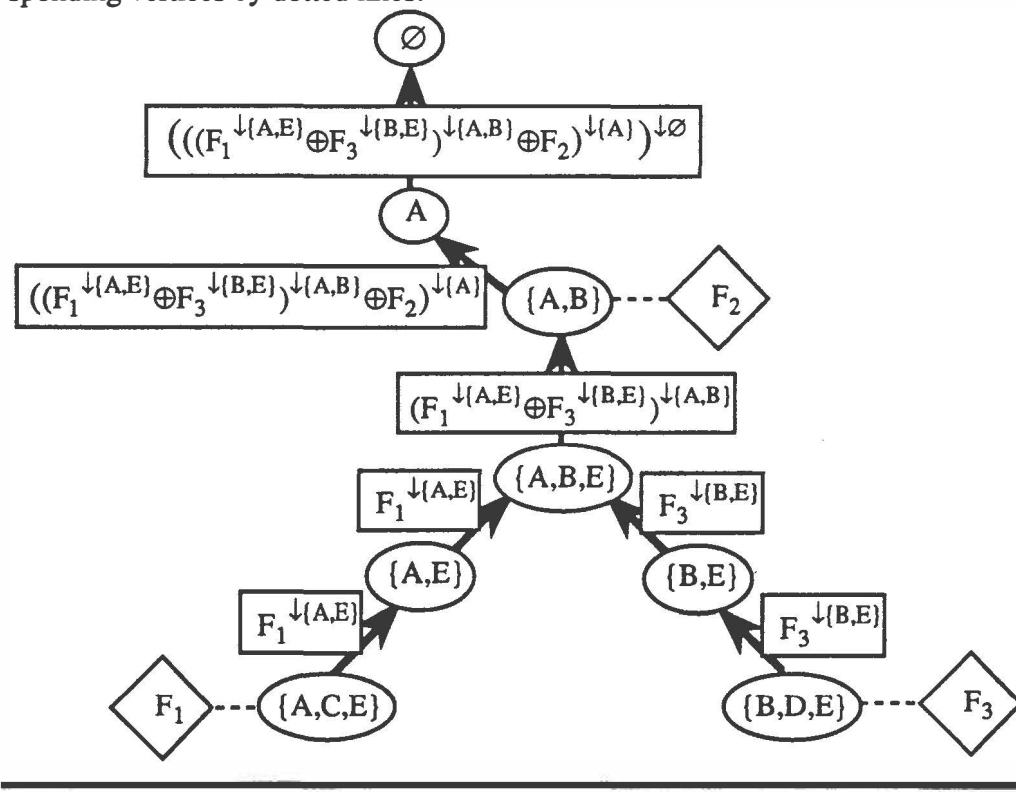

### 4.3. Phase Three: Finding a Solution

In phase two, each time we marginalize a variable, assume that we store the corresponding solution for that variable at the vertex where we do the marginalization. For example, in the optimization problem, we store a solution for C at vertex $\{A,C,E\}$, we store a solution for D at vertex $\{B,D,E\}$, we store a solution for E at vertex $\{A,B,E\}$, we store a solution for B at vertex $\{A,B\}$, and we store a solution for A at vertex $\{A\}$ (see Figures 4, 5, and 6).

In this phase, each vertex of the rooted Markov tree sends a configuration to each of its children. We shall call the configuration transmitted by vertex $h_i$ to its child $h_j \in Ch(h_i)$ as a *con-

*figuration message* and denote it by $c^{h_i \to h_j}$. $c^{h_i \to h_j}$ will always be an element of $\mathcal{W}_{h_i \cap h_j}$. As in phase two, we will give a recursive definition of configuration messages.

The messages start at the root and travel toward the leaves. The configuration message from vertex $\varnothing$ to its child, say $h_1$, is given by

$$c^{\varnothing \to h_1} = \blacklozenge. \tag{4.3}$$

In general, consider vertex $h_i$. It receives a configuration message $c^{Pa(h_i) \to h_i}$ from its parent $Pa(h_i)$. Let $h_j$ be a child of $h_i$. The configuration message from $h_i$ to $h_j$ depends on whether $h_i$ has a solution for a variable stored at its location. (Remember that vertex $h_i$ has a solution for X stored with it if $h_i - Pa(h_i) = \{X\}$).

If $h_i$ has a solution for a variable stored at its location, then

$$c^{h_i \to h_j} = (c^{Pa(h_i) \to h_i}, \Psi_X(c^{Pa(h_i) \to h_i}))^{\downarrow(h_i \cap h_j)} \tag{4.4}$$

where X is such that $\{X\} = h_i - Pa(h_i)$.

If $h_i$ has no solution for a variable stored at its location, then

$$c^{h_i \to h_j} = (c^{Pa(h_i) \to h_i})^{\downarrow(h_i \cap h_j)}. \tag{4.5}$$

We stop the message passing process when each vertex that has a solution stored at its location has received a configuration message.

***Theorem 2.*** Suppose $h_X$ denotes the vertex that has the solution for X stored at its location. Then $z \in \mathcal{W}_{\mathfrak{X}}$ given by

$$z^{\downarrow\{X\}} = \Psi_X(c^{Pa(h_X) \to h_X}) \tag{4.6}$$

for every $X \in \mathfrak{X}$ is a solution for $F_1 \oplus ... \oplus F_k$.

Figure 6 illustrates the message passing scheme for the optimization problem. As per Theorem 2, a solution for F is given by $(\Psi_A(c^{\varnothing \to \{A\}}), \Psi_B(c^{\{A\} \to \{A,B\}}), \Psi_C(c^{\{A,E\} \to \{A,C,E\}}),$ $\Psi_D(c^{\{B,E\} \to \{B,D,E\}}), \Psi_E(c^{\{A,B\} \to \{A,B,E\}}))$. From Figures 5 and 6, it is clear that configurations $(\sim a,b,c,d,e)$ and $(\sim a,b,\sim c,d,e)$ are both optimal for F.

## Acknowledgements

This work was supported in part by the National Science Foundation under grant IRI-8902444. This paper is a revision of Shenoy and Shafer [1988a] that I first wrote in 1988. I am grateful for comments from Judea Pearl, Glenn Shafer and Po-Lung Yu on previous drafts of this paper. Due to page limitations, proofs are omitted. Proofs will be included in a working paper with the same title as this paper.

## References

Arnborg, S., Corneil, D. G. and Proskurowski, A. (1987), "Complexity of finding embeddings in a k-tree," *SIAM Journal of Algebraic and Discrete Methods*, **8**, 277-284.

Bellman, R. E. (1957), *Dynamic Programming*, Princeton University Press, Princeton, NJ.

Bertele, U. and Brioschi, F. (1972), *Nonserial Dynamic Programming*, Academic Press, New York, NY.

Dempster, A. P. and Kong, A. (1988), "Uncertain evidence and artificial analysis," *Journal of Statistical Planning and Inference*, **20**, 355-368.

Kong, A. (1986), "Multivariate belief functions and graphical models," Ph.D. thesis, Department of Statistics, Harvard University, Cambridge, MA.

Lauritzen, S. L. and Spiegelhalter, D. J. (1988), "Local computations with probabilities on graphical structures and their application to expert systems (with discussion)," *Journal of the Royal Statistical Society*, series B, **50**(2), 157-224.

Mellouli, K. (1987), "On the propagation of beliefs in networks using the Dempster-Shafer theory of evidence," Ph.D. thesis, School of Business, University of Kansas, Lawrence, KS.

**Figure 5.** The details of the valuation messages for the optimization problem.

| $\mathcal{W}_{\{A,C,E\}}$ | | | $F_1$ |
|---|---|---|---|
| a | c | e | 1 |
| a | c | ~e | 3 |
| a | ~c | e | 5 |
| a | ~c | ~e | 8 |
| ~a | c | e | 2 |
| ~a | c | ~e | 6 |
| ~a | ~c | e | 2 |
| ~a | ~c | ~e | 4 |

| $\mathcal{W}_{\{A,E\}}$ | | $F_1^{\downarrow(A,E)}$ | $\Psi_C$ |
|---|---|---|---|
| a | e | 1 | c |
| a | ~e | 3 | c |
| ~a | e | 2 | c or ~c |
| ~a | ~e | 4 | ~c |

| $\mathcal{W}_{\{B,D,E\}}$ | | | $F_3$ |
|---|---|---|---|
| b | d | e | 0 |
| b | d | ~e | 5 |
| b | ~d | e | 6 |
| b | ~d | ~e | 3 |
| ~b | d | e | 5 |
| ~b | d | ~e | 1 |
| ~b | ~d | e | 4 |
| ~b | ~d | ~e | 3 |

| $\mathcal{W}_{\{B,E\}}$ | | $F_3^{\downarrow(B,E)}$ | $\Psi_D$ |
|---|---|---|---|
| b | e | 0 | d |
| b | ~e | 3 | ~d |
| ~b | e | 4 | ~d |
| ~b | ~e | 1 | d |

| $\mathcal{W}_{\{A,B,E\}}$ | | | $F_1^{\downarrow(A,E)}$ | $F_3^{\downarrow(B,E)}$ | $F_1^{\downarrow(A,E)}\oplus F_3^{\downarrow(B,E)}$ |
|---|---|---|---|---|---|
| a | b | e | 1 | 0 | 1 |
| a | b | ~e | 3 | 3 | 6 |
| a | ~b | e | 1 | 4 | 5 |
| a | ~b | ~e | 3 | 1 | 4 |
| ~a | b | e | 2 | 0 | 2 |
| ~a | b | ~e | 4 | 3 | 7 |
| ~a | ~b | e | 2 | 4 | 6 |
| ~a | ~b | ~e | 4 | 1 | 5 |

| $\mathcal{W}_{\{A,B\}}$ | | $(F_1^{\downarrow(A,E)}\oplus F_3^{\downarrow(B,E)})^{\downarrow(A,B)}$ | $\Psi_E$ |
|---|---|---|---|
| a | b | 1 | e |
| a | ~b | 4 | ~e |
| ~a | b | 2 | e |
| ~a | ~b | 5 | ~e |

| $\mathcal{W}_{\{A,B\}}$ | | $(F_1^{\downarrow(A,E)}\oplus F_3^{\downarrow(B,E)})^{\downarrow(A,B)}$ | $F_2$ | $(F_1^{\downarrow(A,E)}\oplus F_3^{\downarrow(B,E)})^{\downarrow(A,B)}\oplus F_2$ |
|---|---|---|---|---|
| a | b | 1 | 4 | 5 |
| a | ~b | 4 | 8 | 12 |
| ~a | b | 2 | 0 | 2 |
| ~a | ~b | 5 | 5 | 10 |

| $\mathcal{W}_{\{A\}}$ | $((F_1^{\downarrow(A,E)}\oplus F_3^{\downarrow(B,E)})^{\downarrow(A,B)}\oplus F_2)^{\downarrow(A)}$ | $\Psi_B$ |
|---|---|---|
| a | 5 | b |
| ~a | 2 | b |

| $\mathcal{W}_{\varnothing}$ | $(((F_1^{\downarrow(A,E)}\oplus F_3^{\downarrow(B,E)})^{\downarrow(A,B)}\oplus F_2)^{\downarrow(A)})^{\downarrow\varnothing}$ | $\Psi_A$ |
|---|---|---|
| ♦ | 2 | ~a |

**Figure 6**. The propagation of configuration messages in the optimization problem. The configuration messages are shown as rectangles overlapping the corresponding edge. Note that the direction of messages is opposite to the direction of the edges. The solutions for the five variables are shown as inverted rectangles attached to the vertex (where they are stored) by dotted lines.

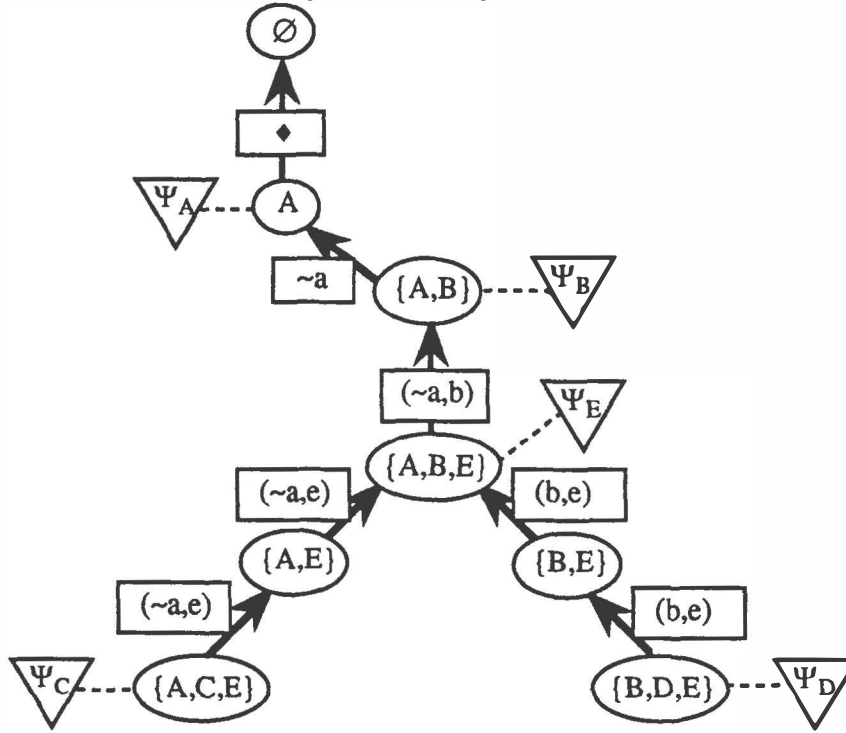

Mitten, L. G. (1964), "Composition principles for synthesis of optimal multistage processes," *Operations Research*, **12**, 610-619.

Pearl, J. (1986), "Fusion, propagation and structuring in belief networks," *Artificial Intelligence*, **29**, 241-288.

Shenoy, P. P. (1989a), "A valuation-based language for expert systems," *International Journal for Approximate Reasoning*, **3**(5), 383-411, 1989.

Shenoy, P. P. (1989b), "On Spohn's rule for revision of beliefs," Working Paper No. 213, School of Business, University of Kansas, Lawrence, KS.

Shenoy, P. P. (1990a), "Valuation-based systems for Bayesian decision analysis," Working Paper No. 220, School of Business, University of Kansas, Lawrence, KS.

Shenoy, P. P. (1990b), "Consistency in valuation-based systems," Working Paper No. 216, School of Business, University of Kansas, Lawrence, KS. To appear in *Methodologies for Intelligent Systems*, **5**, 1990, North-Holland.

Shenoy, P. P. and Shafer, G. (1986), "Propagating belief functions using local computations," *IEEE Expert*, **1**(3), 43-52.

Shenoy, P. P. and Shafer, G. (1988a), "Axioms for discrete optimization using local computation," Working Paper No. 207, School of Business, University of Kansas, Lawrence, KS.

Shenoy, P. P. and Shafer, G. (1988b), "Constraint propagation," Working Paper No. 208, School of Business, University of Kansas, Lawrence, KS.

Zhang, L. (1988), "Studies on finding hypertree covers of hypergraphs," Working Paper No. 198, School of Business, University of Kansas, Lawrence, KS.